\documentclass[pmlr]{jmlr}

\RequirePackage{graphicx}
 \usepackage{booktabs}
\usepackage{longtable}

\usepackage{siunitx}
\usepackage{multirow}
\usepackage{cleveref}

\usepackage[switch]{lineno}

\usepackage{caption}

\makeatletter
\def\set@curr@file#1{\def\@curr@file{#1}} 
\makeatother
\usepackage[load-configurations=version-1]{siunitx} 

\theorembodyfont{\upshape}
\theoremheaderfont{\scshape}
\theorempostheader{:}
\theoremsep{\newline}

\jmlrvolume{298}
\jmlryear{2025}
\jmlrworkshop{Machine Learning for Healthcare}

\title[Embedded Neural Hawkes Process for Diagnostic Trajectories]{Balancing Interpretability and Flexibility in Modeling Diagnostic Trajectories with an Embedded Neural Hawkes Process Model}

\author{%
\Name{Yuankang Zhao} 
\Email{yuankang.zhao@duke.edu}\\
\addr Department of Biostatistics and Bioinformatics\\
       Duke University\\
       Durham, North Carolina, USA 
\AND
\Name{Matthew M. Engelhard}
\Email{m.engelhard@duke.edu}\\
\addr Department of Biostatistics and Bioinformatics\\
       Duke University\\
       Durham, North Carolina, USA 
}

\begin{document}

\maketitle

\begin{abstract}
The Hawkes process (HP) is commonly used to model event sequences with self-reinforcing dynamics, including electronic health records (EHRs). Traditional HPs capture self-reinforcement via parametric impact functions that can be inspected to understand how each event modulates the intensity of others.
Neural network-based HPs offer greater flexibility, resulting in improved fit and prediction performance, but at the cost of interpretability, which is often critical in healthcare.
In this work, we aim to understand and improve upon this tradeoff.
We propose a novel HP formulation in which impact functions are modeled by defining a flexible impact kernel, instantiated as a neural network, in event embedding space, which allows us to model large-scale event sequences with many event types.
This approach is more flexible than traditional HPs
yet more interpretable than other neural network approaches, 
and allows us to explicitly trade flexibility for interpretability by adding transformer encoder layers to further contextualize the event embeddings.
Results show that our method accurately recovers impact functions in simulations, achieves competitive performance on MIMIC-IV procedure dataset, and gains clinically meaningful interpretation on Duke-EHR with children diagnosis dataset  
even without transformer layers. This suggests that our flexible impact kernel is often sufficient to capture self-reinforcing dynamics in EHRs and other data effectively, implying that interpretability can be maintained without loss of performance. \footnote{ Open-source Code for our ENHP model: \url{https://github.com/engelhard-lab/embedded-hp}}
\end{abstract}

\section{Introduction}

The Hawkes process (HP) is a powerful tool for modeling event sequences with self-reinforcing dynamics, making it applicable to domains such as electronic health records (EHRs) \citep{wang2018supervised}, stock trading \citep{bacry2015hawkes}, and social media interactions \citep{yang2011like}. Unlike the Poisson process, which assumes events occur independently over time with a constant intensity, the HP captures the influence of past events on the likelihood of future occurrences by introducing self-excitation. This self-reinforcing property allows the HP to model the clustering behavior frequently observed in real-world event data \citep{hawkes1971spectra, ogata1988statistical}. For instance, in EHRs, a patient’s sequence of diagnoses, prescriptions, and lab tests can reflect how each medical event influences subsequent treatments and health outcomes. While much research focuses on predicting future events \citep{shchur2019intensity,mei2017neural}, we aim to build interpretable models that reveal how events influence one another throughout time, shedding light on Granger causality. \citep{zhang2020cause}

However, capturing self-reinforcement in traditional Hawkes Processes (HPs) often relies on parametric impact functions, which may inadequately represent non-linear dependencies in EHRs and real-world data \citep{linderman2014discovering}. To overcome these limitations, researchers increasingly integrate neural networks with point processes for greater adaptability. For example, Neural Hawkes Processes use recurrent neural networks to effectively model non-linear, long-range temporal dependencies \citep{mei2017neural}, while Transformer Hawkes Processes apply self-attention mechanisms to manage variable-length event dependencies \citep{zuo2020transformer}. Additionally, Intensity-Free Temporal Point Processes directly model inter-event time distributions, circumventing computationally intensive integral calculations of traditional intensity functions \citep{shchur2019intensity}. Neural Spatio-Temporal Point Processes (ODETPP) leverage neural ordinary differential equations to model temporal dependencies more simply, omitting spatial components \citep{chen2020neural}. While these methods enhance intensity function representation, they typically reduce interpretability \citep{xu2020learning}.

Many of these methods, while powerful, relax so many of the Hawkes process’s original assumptions that they lose key features and become black-box models \citep{linderman2014discovering}. This lack of transparency makes it difficult to interpret event interactions or uncover Granger causality, which is critical in healthcare \citep{eichler2017graphical}. In healthcare, understanding the relationships between medical events is as important as predicting the events themselves. For example, identifying how medical events relate can guide clinical research and improve patient care \citep{liu2019early}, and emerging regulations governing the use of models for clinical decision-making emphasize interpretability \citep{FDA2023}. To serve these needs, we require models that balance flexibility with transparency.

In this work, we aim to understand and improve the tradeoff between interpretability and flexibility in HP models. To do this, we propose a novel HP formulation in which impact functions are modeled by defining a flexible, neural network-based impact kernel in event embedding space. Our formulation maintains the core formulation of the Hawkes process, combining a baseline intensity with an impact function summed over all previous events, and thereby retains the Hawkes process’s key properties, such as positive intensity and additive influence. However, to improve flexibility, we replace the traditional exponential decay assumption \citep{hawkes1971spectra} with neural network-driven impact functions, allowing for more nuanced modeling of event dependencies without sacrificing interpretability. 

By working in embedding space, we limit the dimensionality of our impact kernel, which allows us to model large-scale event sequences with tens of thousands of event types. Each dimension in this space represents a broader event topic, with the impact kernel capturing the relationships between these topics. This makes large-scale event modeling computationally feasible while enhancing interpretability, as the interactions are understood at the topic level.

In some cases, it may be desirable to trade interpretability for greater flexibility. Our approach allows users to explore this tradeoff directly by adding optional transformer encoder layers to contextualize the embeddings of each event in a given sequence based on the previous history of events. Adding these layers reduces interpretability and sacrifices the property of additive influence, but in principle, it also allows the model to capture more complex dependencies between event types. Importantly, however, our results show that this sacrifice is rarely necessary, because our flexible impact kernel alone is sufficient to capture the dynamics of most real-world sequences.

In summary, our contributions are:

\begin{enumerate}
    \item \textbf{Novel Hawkes Process Formulation:} We introduce a generalized Hawkes process where impact functions are defined via a flexible, neural network-based impact kernel within an event embedding space.
    \item \textbf{Controlled Tradeoff Between Interpretability and Flexibility:} We develop a method to explicitly manage the balance between interpretability and model complexity by incorporating transformer encoder layers to contextualize event embeddings based on the historical sequence of events.
    \item \textbf{Maintaining Interpretability Without Sacrificing Performance:} We demonstrate that in real-world settings, transformer encoder layers are often unnecessary to achieve state-of-the-art performance, thereby maintaining interpretability without compromising the model's effectiveness.

    \item \textbf{Application to EHR Data:} To demonstrate utility, we apply our method to a pediatric EHR dataset. Our model supports clinical interpretation in two ways. First, by learning event embeddings and a topic-level impact kernel, it identifies links between medical event categories, e.g., how perinatal complications affect later neurodevelopmental conditions. Second, given specific clinical hypotheses—such as the influence of speech delay or early behavioral disorders on ADHD—our model estimates temporal impact functions $\phi_{i,j}(t)$. Similar to traditional Hawkes processes but unlike recent neural methods \citep{zuo2020transformer,boyd2023inference,zhang2024neural}, this reveals evolving temporal influences, highlighting critical intervention periods. Thus, our approach effectively supports clinical hypothesis generation and validation.
    \end{enumerate}

 \paragraph{Generalizable Insights about Machine Learning in the Context of Healthcare:} Understanding how diagnoses and other EHR events mutually influence one another is important to understand patient trajectories and consider ways in which the patient journey might be optimized or otherwise improved. However, it is difficult to design a traditional inferential analysis to understand the influence of some events on others, because there are so many different event types, and a given event can occur repeatedly and at any time. Hawkes process models are ideally suited to this task, but existing methods are either (a) insufficiently flexible to fully capture the temporal dynamics of EHR sequences, or (b) insufficiently flexible to provide clear understanding or takeaways.
Our framework strikes a balance between flexibility and interpretability by learning impact functions in a structured, low-dimensional embedding space while preserving key properties of the Hawkes process (e.g., positive, additive influence). This approach offers a broadly applicable strategy for uncovering temporal dependencies in healthcare data, enabling domain experts to formulate and validate hypotheses across diverse clinical domains. Beyond the specific method we develop, we believe this work illustrates benefits of using a Hawkes process approach to understand diagnostic trajectories.

\section{Related work} 
\subsection{Generalized Hawkes Process}
Traditional Hawkes processes are limited in their ability to model complex event dynamics due to their reliance on a simple exponential decay for the event intensity. To address these limitations, neural network-based extensions have been developed. One key advancement is the Neural Hawkes Process(NHP) (\cite{mei2017neural}), which generates events sequentially using recurrent neural networks. NHP replaces the traditional intensity function with one parameterized by a recurrent model, allowing it to better capture dependencies between events. 

Another generalized Hawkes process is the attention-based Hawkes process. One of the classic example is the Transformer Hawkes Process(THP)\citep{zuo2020transformer}. THP leverages attention mechanisms to model event intensity, using transformer-based encodings to incorporate both temporal and contextual information. It enhances the intensity function by introducing terms to account for event timings and baseline effects. By doing so, THP captures the influence of past events on future ones in a more flexible and context-aware manner. Like other generalized Hawkes processes, both NHP and THP are trained to maximize the likelihood of observed event sequences, enabling them to model more complex event interactions.

\subsection{Self-Attentive Hawkes Process(SAHP)}
The Self-Attentive Hawkes Process(SAHP) \citep{zhang2020self} extends the traditional Hawkes process by incorporating a self-attention mechanism. This enhancement allows for better modeling of complex event dependencies and improves interpretability. SAHP quantifies the influence of historical events on future ones using attention weights, providing an interpretable measure of how past events affect subsequent occurrences. By accumulating these attention weights, SAHP effectively quantifies the statistical influence between different event types, making it valuable for both predicting event sequences and explaining event relationships.

However, SAHP represents a significant departure from the classical Hawkes process, which models the impact of events through explicit time-decaying influence functions. In contrast, SAHP relies on learned attention weights without explicitly capturing the temporal dynamics of influence decay inherent in the original impact function. In contrast, our method retains the explicit time-decay structure by parameterizing the impact function with a neural network, thereby preserving the interpretable temporal dynamics while also allowing for greater flexibility in modeling complex event interactions.

\section{Methods}

\subsection{The Hawkes Process}

Let us denote an event sequence as \( \mathcal{S} = \{(t_1, k_1), (t_2, k_2), \ldots, (t_L, k_L)\} \), where $L$ is the number of events in the sequence, $k_i \in \{1,2...,M\}$ is the event type of the $i$th event, and \( t_i \) is the time of $i$th event. For each event type \( k \), a counting process \( N_k(t) \) records the cumulative number of events that have occurred up until time \( t \). The intensity function \( \lambda_k(t) \) is defined as the expected instantaneous rate of type-\( k \) events given the history of events, formalized as:

\[
\lambda_k(t) = \frac{E[dN_k(t) | \mathcal{H}_t]}{dt}, \quad \mathcal{H}_t = \{(t_i, k_i) | t_i < t, k_i \in \{1 \dots M\}\}
\]

In a standard Hawkes process, which is a type of self-exciting multivariate point process, the intensity depends on the history of past events. The intensity function is defined as:

\begin{align}
\label{eq:intensity_hp}
\lambda_k(t) &= \mu_k + \sum_{k'=1}^{M} \int_0^t \phi_{k',k}(t-s) dN_{k'}(s) \\  
&= \mu_k + \sum_{(t_i, k_i) \in \mathcal{H}_t} \alpha_{k_i,k} \exp(-\delta_{k_i,k} (t - t_i))
\end{align}
In this standard Hawkes processes, the impact functions $\phi_{k',k}(t-s)$ are assumed to follow an exponential decay. This intensity function provides insight into how likely it is for a specific event type to occur at any given moment in time, considering the past events up to time \( t \).

Here, \( \mu_k \) is the baseline intensity, and \( \phi_{k',k}(t-s) \) is the impact function that quantifies how events of type \(k'\) affect the intensity of events of type \(k\). \( \alpha_{k_i,k} \) controls the strength of the triggering effect, while \( \delta_{k_i,k} \) determines how fast the influence decays over time.

\subsection{Impact Kernel Sub-Network}
In this work, we aim to relax the assumption that the impact functions follow a particular parametric form by introducing a neural network-based impact function. We begin by simplifying \eqref{eq:intensity_hp} for $t_j \in (0, T)$, where $T$ is the maximum observation time. The intensity function for the $k_i$-th event, $\lambda_{k_i}(t_j)$, is expressed as:

\begin{equation}
\label{eq:intensity}
    \lambda_{k_i}(t_j) = \mu_{k_i} + \sum_{i<j} \phi_{k_i,k_j}(t_j - t_i)
\end{equation}

where $\mu_{k_i}$ is the base intensity for event type $k_i$, and $\phi_{k_i, k_j}(\Delta t)$ represents the \textit{impact function} that quantifies the influence of an event of type $k_i$ occurring at time $t_i$ on the intensity $\lambda_{k_i}(t_j)$ at a later time $t_j$. This effect depends solely on the time difference $\Delta t = t_j - t_i$.

The total intensity $\lambda(t)$ for any event occurring at time $t$ is given by the sum of intensities over all event types, i.e., $\lambda(t) = \sum_{k=1}^M \lambda_k(t)$.

In our proposed approach, the impact functions $\phi_{i,j}(\Delta t)$ are modeled using an impact kernel  $K(\Delta t)$  with  $M^2$  outputs, where  M  is the number of event types. The kernel  $K(\Delta t)$  is parameterized by a neural network that takes $\Delta t$ as input and outputs the impact relationships among all event type pairs. As shown in Figure \ref{fig:method}, without event embedding, the impact function for each pair of event types is obtained by selecting the corresponding elements from the output of  $K(\Delta t) $. The impact function can be written as:
\begin{equation}
\phi_{i,j}(\Delta t) = (e^{(i)})^\top K(\Delta t) e^{(j)}
\label{eq:method_eij}
\end{equation}
Here,  $K(\Delta t)$  encompasses all the impact kernels between event type pairs $(i, j)$ for  $i, j \in \{1,2... M\}$ , and $ e^{(i)}$  and  $e^{(j)}$  are the corresponding one-hot vectors of the event types at times  $t_i$  and  $t_j$ , respectively. The neural network used to parameterize $ K(\Delta t)$  can have a simple or complex architecture. Although it is parameterized, we do not make any assumptions about the specific shape of the impact functions. Our experiments demonstrate that even simple neural networks are sufficient for modeling impact functions in real-world data. As the number of event types increases, the computational cost of modeling impact functions grows quadratically. However, if the $M \dot M$ kernel has a sparse structure, most event pairs have no impact to each other, and modeling all pairs becomes unnecessary. In such cases, embedding-based methods are more efficient, as discussed in the next section.

To optimize the model, we aim to maximize the log-likelihood  $\ell(\mathcal{S})$  of an event sequence  $\mathcal{S}$ , which is given as follows \citep{mei2017neural}:

\begin{equation}
\label{eq:loglik}
    \ell(\mathcal{S}) = \sum_{i=1}^L \log \lambda_{k_i}(t_i) - \int_0^T \lambda(t) dt
\end{equation}

The first term in \eqref{eq:loglik} involves evaluating the impact function $\phi$ over $\mathcal{O}(L^2)$ time intervals. The second term, an integral, can be approximated using numerical methods or Monte Carlo integration. Although numerical methods may introduce bias depending on the approach, they tend to outperform Monte Carlo methods due to the latter's high variance. The numerical approximation can be represented as linear interpolation between observed events:

\begin{align}
\label{eq:non-event}
\int_0^T \lambda(t) dt &\approx \sum_{j=2}^L \frac{(\lambda(t_j) + \lambda(t_{j-1}))(t_j - t_{j-1})}{2}
\end{align}

Alternatively, Monte Carlo integration estimates the integral as:

\begin{equation}
\label{eq:non-event-MC}
\int_0^T \lambda(t) dt  \approx \sum_{j=2}^L \left( \frac{1}{N} \sum_{i=1}^N \lambda(t_{j-1,i}) \right)(t_j - t_{j-1})
\end{equation}

where $t_{j-1,i} \sim \text{Unif}(t_{j-1}, t_j)$ and $N$ is the number of samples drawn. Monte Carlo methods require $\mathcal{O}(L^2NM)$ evaluations of $\phi$, while numerical methods require $\mathcal{O}(L^2K)$, making the latter more efficient for large datasets.

\subsection{Formulating the Impact Kernel in Embedding Space}

Let \( W \in \mathbb{R}^{D \times M} \) represent the embedding matrix, where \( M \) is the number of event types and \( D \) is the embedding dimension. The embedding for event type \( i \) is given by \( W e^{(i)} \), where \( e^{(i)} \) is the standard basis vector selecting the \( i \)-th column of \( W \). Similarly, the embedding for event type \( j \) is \( W e^{(j)} \). The embedding matrix \( W \) may be shared between the input and output events or may differ, i.e., \( W_1 e^{(i)} \) for input and \( W_2 e^{(j)} \) for output events.

For a given time interval between events, we define an impact kernel \( K(\Delta t) \in \mathbb{R}^{D \times D} \) in the embedding space. In this formulation, the impact kernel is instantiated as a neural network with \( D^2 \) outputs. The impact of an event of type \( i \) at time \( t_i \) on the intensity of an event of type \( j \) at time \( t_j \) is expressed as:

\begin{equation}
    \phi_{i, j}(t_j - t_i) = (W e^{(i)})^\top K(t_j - t_i) (W e^{(j)})
    \label{eq:method_embed}
\end{equation}

Intuitively, for each event pair \( (k_i, k_j) \), we select columns from the embedding matrix \( W \) and compute a linear combination of the impact kernels, weighted by each event's contribution, to reconstruct the impact function for that pair. This method, referred to as the Embedded Neural Hawkes Process(ENHP), reduces the dimensionality of the impact kernels, making it well-suited for high-dimensional event spaces, while also capturing the sparse structure of event interactions. Additionally, the embedding matrix \( W \) introduces interpretability by mapping events to latent topics, where each dimension corresponds to a distinct attribute. To ensure the non-negativity of the intensity, we apply a softplus function to all components, including \( W \), \( K(t) \), and \( \mu_k \).

\begin{figure}[t]
\floatconts
    {fig:method}
    {\caption{Architecture of the ENHP and ENHP-C. Without the event embedding and transformer encoder, the impact function follows formula\ref{eq:method_eij}. Without the transformer encoder, it follows formula\ref{eq:method_embed}. With all components, it corresponds to formula\ref{eq:method_TF}.}}
    {\includegraphics[width=3in]{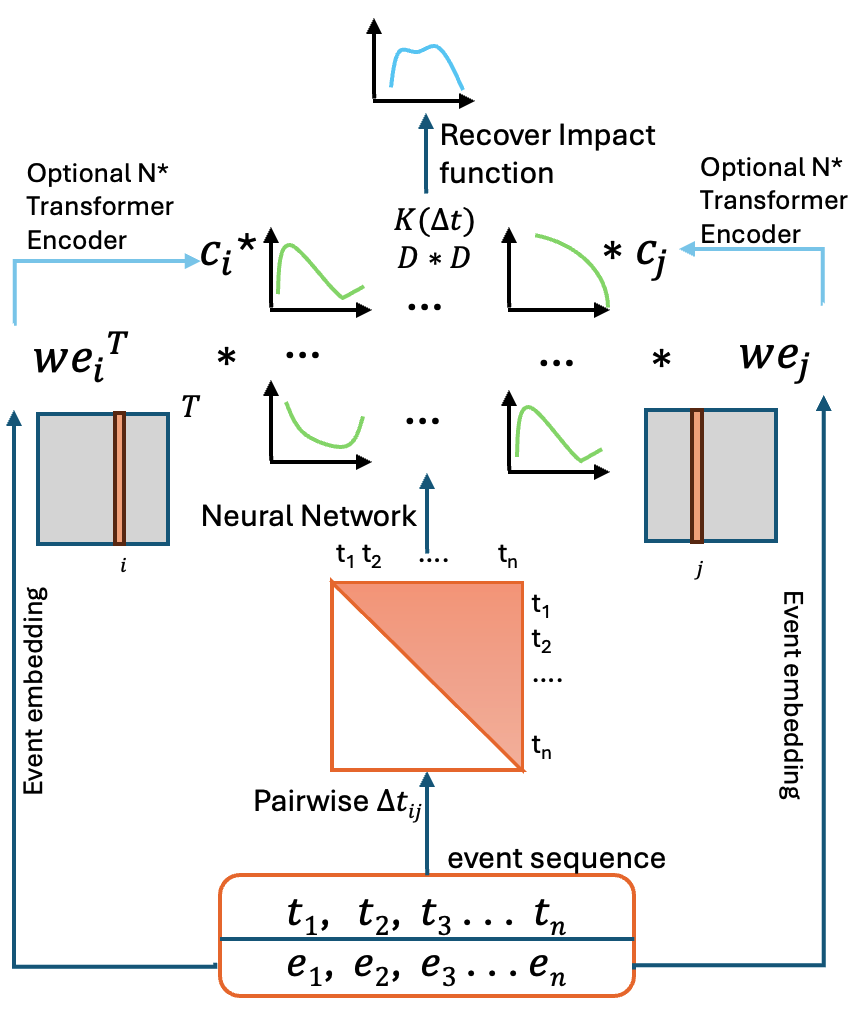}}
\end{figure}
We further enhance the model by introducing a transformer encoder on top of the event embeddings to capture contextualized event interactions. The contextualized embeddings generated by the transformer are given by:

\begin{equation}
\phi_{i,j}(t_j - t_i) = (c^{(i)})^\top K(t_j - t_i) (c^{(j)})
\label{eq:method_TF}
\end{equation}

where  $c^{(i)}$  represents the output of the transformer encoder, which takes the sum of the event embedding and a temporal embedding as input. This extended model is termed the Embedded Neural Hawkes Process - Contextualized (ENHP-C). While this approach can potentially enhance performance by modeling more intricate event relationships and improving likelihood, the learned kernel impact functions become less interpretable, as the contextualized embeddings are hidden representations.

In this paper,  $K(\Delta t)$  is parameterized by a neural network consisting of a fully connected layer with ReLU activation, followed by a linear output layer.

\section{Experiments}

\subsection{Baseline}
All the intensity-based generalized Hawkes process models mentioned in the related work section will be included as baseline methods, including NHP, THP, and SAHP, which are introduced in the related work section. Additionally, we include the following method:

\paragraph{Attentive Neural Hawkes Process:}
AttNHP \citep{yang2022transformer} introduces an attention mechanism into the traditional Neural Hawkes Process to enhance its ability to model event dependencies. In this approach, instead of relying solely on a recurrent neural network to capture the influence of past events, attention weights are applied to determine the relative importance of each historical event in predicting future occurrences. This makes the model more accurate and easier to understand because it focuses on the key events in the history.

\subsection{Real-world Datasets}
To evaluate our model, we utilize several well-established datasets from different domains. Each dataset represents a sequence of events with time stamps and categorical labels defining the event types. With the exception of the MIMIC-IV dataset, results for baseline methods an all datasets are reproduced from previously published work \citep{mei2017neural, xue2023easytpp}. The descriptions of the datasets used in our experiments are in appendix B.

\subsection{Simulation}
\begin{figure}[h]
    \centering
\includegraphics[width=0.8\columnwidth]{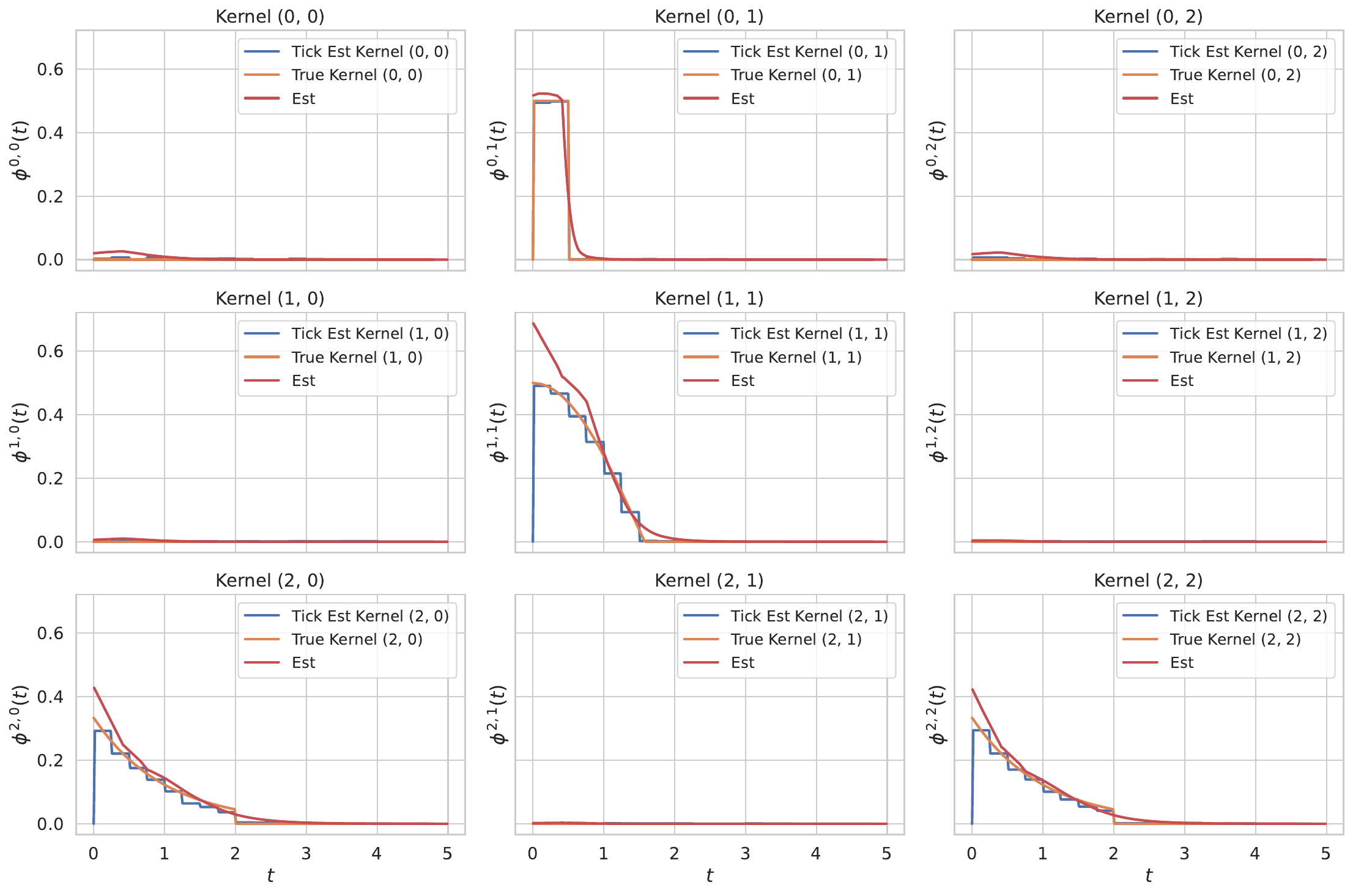}
\caption{Fitted triggering kernel using ENHP}
\label{fig:simu_embed}
\end{figure}
We begin by verifying that our model successfully fits data generated from a known distribution, and that the impact kernels learned by our method match the known, true impact kernels. We generate a synthetic dataset using \emph{tick}, an open-source machine learning library for Python that includes a Hawkes process module \citep{JMLR:v18:17-381}. Specifically, a three-dimensional Hawkes process was generated with baseline intensities $\mu_0=0.3$, $\mu_1=0.05$ and $\mu_2=0.2$. The triggering kernels include four active (\textit{i.e.}, nonzero) kernels: a step function, a cosine kernel, and two exponential kernels.  Other kernels are inactive (i.e., identically zero). The details are as follows: $\phi_{0,1}(t) = 0.5$ for $t \in [0, 0.5]$ and $\phi_{0,1}(t) = 0$ otherwise; $\phi_{1,1}(t) = \cos(t/2)$; $\phi_{2,0}(t) = e^{-t/3}$; and $\phi_{2,2}(t) = e^{-t/3}$.

The results are presented in Figure \ref{fig:simu_embed}. Using the proposed ENHP, our method successfully recovers all four active impact functions, regardless of whether the function includes a jump point or is convex or concave. With the exception of the step function kernel, our approach outperforms the estimation from the tick library by providing more accurate shapes for the impact functions. In contrast, tick relies on step functions to approximate impact kernels, which can obscure the true form of the underlying functions. We did observe slight overestimation of the bump in the step function kernel, as well as minor overestimation in the inactive (constant-zero) kernel, but these could be improved by introducing regularization to the embedding matrix. Overall, our method proves effective in recovering various types of impact kernels in a multivariate Hawkes process.

\subsection{Performance on Benchmark Datasets}
\paragraph{Training detail and evaluation protocol} 
We use the EasyTPP open benchmark \citep{xue2023easytpp} to evaluate other methods in our performance comparison. EasyTPP is a newly developed benchmark framework for evaluating temporal point process models. It provides standardized datasets, evaluation metrics, and implementations of various baseline models, enabling fair and consistent comparisons across methods. All hyperparameters for benchmark methods are provided by EasyTPP. For our method, all hyperparameters are included in the appendix. All likelihoods presented below are reported on the validation set. A single NVIDIA RTX A5000 graphics card was used to run all experiments. 

Similar to EasyTPP, we evaluate the models under two standard scenarios: \begin{itemize}
    \item Goodness-of-fit: The models are trained on the training set, and their performance is assessed by measuring the log-likelihood on held-out data.
    \item Next-event prediction: Following the minimum Bayes risk (MBR) framework, we predict the timing of the next event based solely on the preceding event history, and its type using both the observed event time and the preceding event history. The accuracy of time and type predictions is evaluated using Root Mean Square Error(RMSE) and error rate, respectively.
\end{itemize}

\paragraph{Likelihood comparison between proposed methods}
\begin{table}[t]
\centering
\caption{Performance comparison between ENHP and ENHP-C across datasets, with results presented as mean LL ± standard deviation.}
\begin{tabular}{|l|c|c|}
\hline
Data & ENHP & ENHP-C\\ \hline
Retweet      & -3.67 ± 0.02    & -3.80 ± 0.02 \\ 
Amazon       & -0.82 ± 0.03    & -0.82 ± 0.03 \\ 
Taxi         & -0.24 ± 0.04    & -0.23 ± 0.05 \\ 
StackOverflow& -2.43 ± 0.07    & -2.36 ± 0.07 \\ 
MemeTrack    & -10.97 ± 0.11   & -11.45 ± 0.12 \\ 
MIMIC-IV     & -9.61 ± 0.06    & -9.64 ± 0.04 \\
Simulation   & -1.80 ± 0.01    & -1.80 ± 0.01 \\ \hline
\end{tabular}

\label{tab:EHP_comparison}
\end{table}

In this experiment, we compare the performance of the ENHP and ENHP-C in terms of log-likelihood (LL). From Table \ref{tab:EHP_comparison}, we see that the two methods perform similarly across most datasets, with neither model consistently outperforming the other. A contradictory trend is observed: the simpler ENHP model is slightly better on Retweet (3 event types), while the more complex ENHP-C model is slightly better on StackOverflow (22 event types). We hypothesize that the benefit of ENHP-C's additional capacity scales with the dimensionality of the event space. However, since these performance differences are minor, this result demonstrates that our simpler, embedding-based model is already flexible enough to capture the necessary patterns in the data—thus achieving good model fitting while retaining a clear, interpretable structure. In other words, we obtain the benefits of interpretability without sacrificing flexibility. In addition, both models are also capable of handling datasets with a large number of event types, such as the MemeTrack dataset, which contains 5,000 event types. Overall, while contextualizing embeddings using transformer layers may allow the model to capture more complex interactions, the additional benefit is not clearly reflected in the LL results on these real datasets. Given the large size of many of these datasets, we believe it is unlikely that this is the result of insufficient data, and more likely that ENHP is already sufficiently flexible to capture the underlying data distribution.

\paragraph{Likelihood Comparison between Our Method and Other Models}

\begin{figure}
    \centering
    \includegraphics[width=\linewidth]{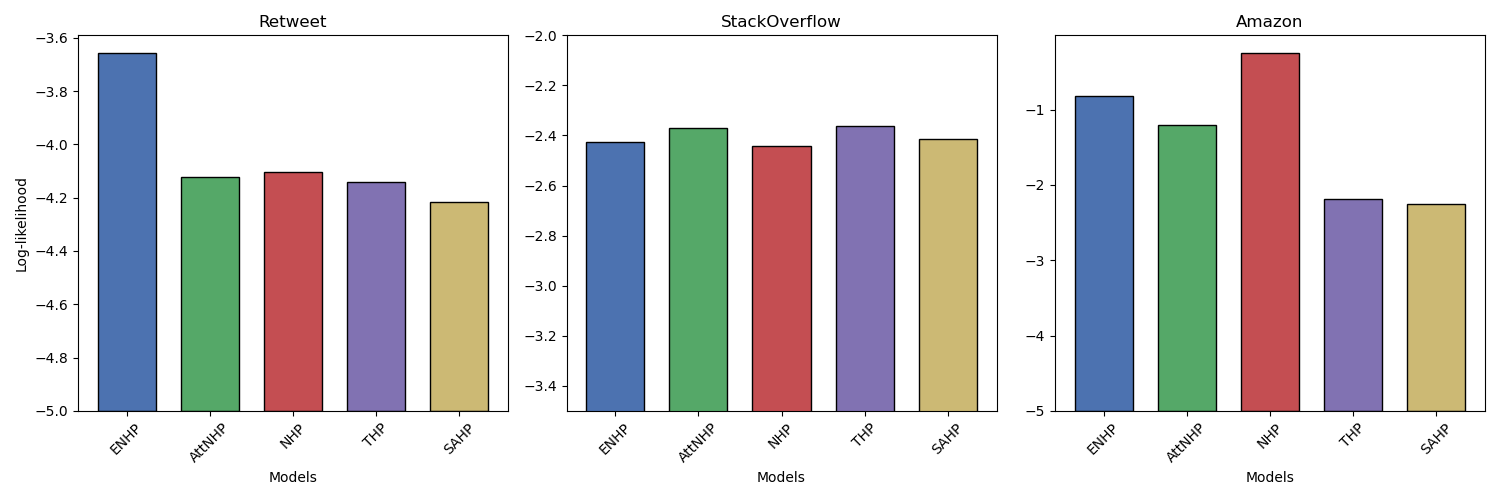}
    \caption{Performance comparison across datasets and models based on log-likelihood}
    \label{fig:LL_fig}
\end{figure}
As shown in figure \ref{fig:LL_fig}, our method demonstrates superior performance across multiple datasets. Specifically, on the Retweet dataset, ENHP achieves the best LL of -3.665, ranking first among all models. On the StackOverflow dataset, All the methods are performed closely with LL around -2.4. For the amazon dataset, ENHP ranks second with LL values of -0.817 narrowly trailing the top-performing model NHP.

Similar patterns have been observed in previous studies \citep{xue2023easytpp, yang2022transformer}, where transformer-based approaches (e.g., THP) do not consistently outperform other attention-based or neural network-based models. This aligns with our findings, where THP does not show significant performance improvements over other models. Additionally, while AttNHP outperforms NHP on the StackOverflow dataset—corroborating the conclusions presented in the AttNHP paper \citep{yang2022transformer}—it performs worse than NHP on other datasets, consistent with results reported in the easyTPP paper for Retweet. Our experimental results are in agreement with these findings, further validating the result of our experiment.

\paragraph{Next-event Prediction comparison between Our Method and Other Models}

\begin{table*}[ht]
\centering
\caption{Next-event Prediction comparison between different models across datasets }
\resizebox{\textwidth}{!}{\begin{tabular}{l|cc|cc|cc|cc|cc}
\hline
\multirow{2}{*}{\textbf{Dataset}} 
 & \multicolumn{2}{c|}{\textbf{ENHP}} 
 & \multicolumn{2}{c|}{\textbf{AttNHP}} 
 & \multicolumn{2}{c|}{\textbf{NHP}} 
 & \multicolumn{2}{c|}{\textbf{THP}} 
 & \multicolumn{2}{c}{\textbf{SAHP}} \\
\cline{2-11}
 & \textbf{RMSE} & \textbf{error} 
 & \textbf{RMSE} & \textbf{error} 
 & \textbf{RMSE} & \textbf{error} 
 & \textbf{RMSE} & \textbf{error} 
 & \textbf{RMSE} & \textbf{error} \\
\hline
\textbf{Retweet} 
 & \textbf{20.204} & 0.414 
 & 22.154 & 0.402 
 & 22.472 & \textbf{0.399} 
 & 23.860 & 0.402 
 & 20.818 & 0.425 \\
\textbf{Taxi} 
 & \textbf{0.331} & 0.104 
 & 0.389 & 0.189 
 & 0.391 & \textbf{0.097} 
 & 0.364 & 0.114 
 & 0.409 & 0.111 \\
\textbf{StackOverflow} 
 & \textbf{1.190} & \textbf{0.519} 
 & 1.354 & 0.537 
 & 1.429 & 0.533 
 & 1.408 & 0.531 
 & 1.319 & 0.561 \\
\textbf{Amazon} 
 & 0.519 & \textbf{0.641} 
 & 0.791 & 0.698 
 & 0.500 & 0.691 
 & \textbf{0.483} & 0.664 
 & 0.539 & 0.688 \\
\textbf{MIMIC-IV} 
 & \textbf{1028.3} & 0.798 
 & 1040.6 & 0.800 
 & 1171.3 & \textbf{0.769} 
 & NA & NA 
 & 1105.1 & 0.789 \\
\hline
\end{tabular}}
\label{tab:PredictionComp}
\end{table*}
Table \ref{tab:PredictionComp} presents a comprehensive comparison of RMSE and error rates of across all benchmark dataset. Our proposed method, ENHP, consistently achieves the lowest RMSE in four out of the five datasets, and rank second on Amazon dataset demonstrating superior predictive accuracy and reliability. For error rate of next event type, ENHP also rank first on StackOverflow and Amazon datasets. For the other three datasets, NHP shows the best performance; however, it is evident that the proposed ENHP demonstrates only a slight difference compared to NHP.  

Note that, except for MIMIC-IV, the other datasets utilize benchmark datasets provided by EasyTPP. EasyTPP typically selects the most active users from the original datasets and removes extreme values in time intervals. In contrast, the MIMIC-IV data is sourced directly from a database, thereby better representing real-world data and posing a challenge to the robustness of the methods. Due to the limitation of EasyTPP version THP, it is unable to process very large time intervals, resulting in the absence of prediction results. On the other hand, this also indicates that our proposed method, ENHP, demonstrates competitive prediction performance on MIMIC-IV and underscores the robustness of our approach in handling extreme values.

Overall, the results indicate that ENHP excels in delivering low RMSE and competitive error rates, affirming its capability as a highly effective model for accurate and dependable predictions across diverse applications.

\subsection{Interpretation of recovering impact function}

\begin{figure*}[htbp]
\floatconts
  {fig:overall} %
  {\caption{Heatmap of different datasets with regard to impact function or kernel function.}} %
  {%
    \subfigure[Amazon impact function]{\label{fig:sub1}%
      \includegraphics[width=0.36\linewidth]{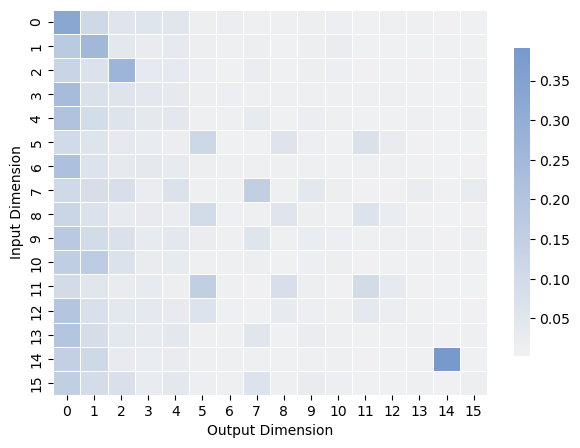}}%
    \hfill
    \subfigure[Stack Overflow impact function]{\label{fig:sub2}%
      \includegraphics[width=0.36\linewidth]{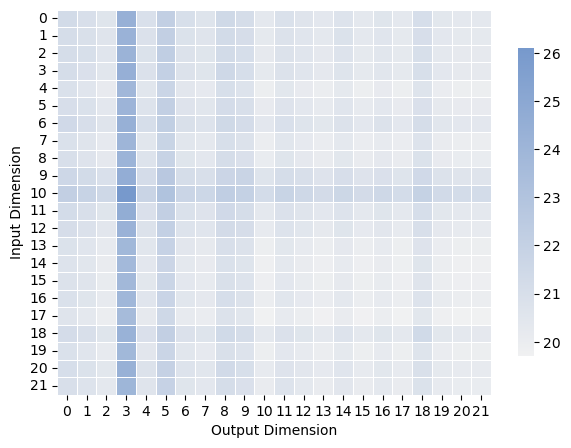}}%
    \hfill
    \subfigure[Duke-EHR Kernel]{\label{fig:sub3}%
      \includegraphics[width=0.265\linewidth]{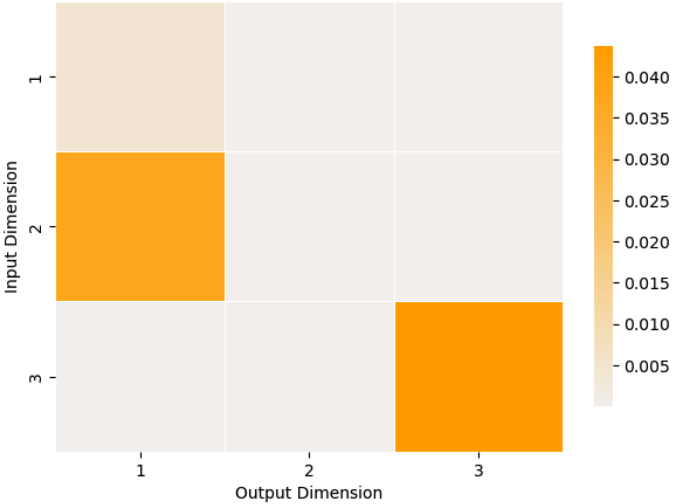}}%
  }
\end{figure*}

As demonstrated in the simulation, our method exhibits strong capability in recovering the impact functions between pairs of events. While it is possible to recover the entire impact function (i.e., intensity over time), this approach is not ideal for visualization due to the high dimensionality and sparse structure of most datasets. Therefore, we compute $\int\phi_{i,j}(t)dt$ for each impact function, which represents the cumulative impact from event $i$ to event $j$. This scalar value is then visualized through a heatmap.
\paragraph{Amazon} Figure \cref{fig:sub1} visualizes the learned impact function for the Amazon dataset, illustrating the influence between different product categories. Several key patterns emerge from this visualization: (1) Self-Excitation on the Diagonal: The darker squares along the diagonal indicate that several categories exhibit strong self-excitation. Notably, categories such as Type 14 (surf, skate, and street), Type 0 (clothing), Type 1 (shoes), and Type 2 (accessories) display high self-excitement. This means that customers are likely to purchase multiple items within the same category in a short period. For Type 14, it is common for customers buying surf or skate gear to need multiple specialized items, which explains the strongest self-excitation in this categories. Similarly, purchasing multiple pieces of clothing, shoes, or accessories in a single session aligns with typical consumer behavior patterns on Amazon, where shoppers often have extra saving for a regular subscription. (2)	The first column, which corresponds to Type 0 (clothing), is notably darker across multiple rows. This indicates that purchases in other categories have a significant impact on clothing purchases. This makes sense given that clothing is a fundamental and prevalent category on this dataset and also in real Amazon sales. 
\paragraph{Stack Overflow}
Figure \cref{fig:sub2} depicts the interactions between different events in the Stack Overflow dataset, where each event represents the awarding of a specific badge to a user. The results reveal that nearly all events tend to trigger Event 3 (Popular Question), followed by Event 5 (Nice Answer). The primary source of excitation is Event 10 (Notable Question), which has a strong influence on other events. A notable observation is that both Event 3 and Event 10 are the most frequent events in this dataset, which aligns with the behavior captured in the data. Our model successfully identifies these patterns, which not only match the known data facts but also reveal new insights into the influence structure within the event space. This suggests that our model is capable of uncovering meaningful and interpretable relationships between events.

\section{Duke EHR case study}

\begin{table*}[!t]
\floatconts
  {tab:topics} 
  {\caption{Event Embedding for Topic Discovery on Duke-EHR}} 
  {\footnotesize 
    \begin{tabular}{p{4.5cm}p{4.5cm}p{4.5cm}} 
      \toprule
      \multicolumn{3}{c}{\textbf{Event Embedding for Input Events}} \\ 
      \midrule
      \textbf{Topic 1} & \textbf{Topic 2} & \textbf{Topic 3} \\ 
      \midrule
      Neonatal jaundice from other and unspecified causes & Acne & Dependence on enabling machines and devices, NEC \\
      Feeding problems of newborn & Newborn affected by maternal complications of pregnancy & Respiratory distress of newborn \\
      Unspecified jaundice & NB aff by matern cond that may be unrelated to present preg & Encounter for attention to artificial openings \\
      Other conditions of integument specific to newborn & Other congenital infectious and parasitic diseases & Cardiovascular disorders originating in the perinatal period \\
      Umbilical hemorrhage of newborn & Newborn affected by other comp of labor and delivery & Transitory disord of carbohydrate metab specific to newborn \\
      \\
      \midrule
      \multicolumn{3}{c}{\textbf{Event Embedding for Output Events}} \\ 
      \midrule
      \textbf{Topic 1} & \textbf{Topic 2} & \textbf{Topic 3} \\ 
      \midrule
      Suppurative and unspecified otitis media & Encounter for immunization & Specific developmental disorders of speech and language \\
      Acute upper resp infections of multiple and unsp sites & Suppurative and unspecified otitis media & Disord of NB related to short gest and low birth weight, NEC \\
      Fever of other and unknown origin  & Persons encntr hlth serv for spec proc \& trtmt, not crd out & Personal risk factors, not elsewhere classified \\
      Cough & Fever of other and unknown origin & Lack of expected normal physiol dev in childhood and adults \\
      Atopic dermatitis & Contact w and (suspected) exposure to communicable diseases &  Attention-deficit hyperactivity disorders \\
      \bottomrule
    \end{tabular}
  }
  \label{table:topics}
\end{table*}
We apply our method to EHR data to illustrate how our method facilitates interpretation of real-world event sequences. This dataset is a retrospective EHR from the Duke University Health System, containing all recorded diagnoses for children under the age of 10. Our analytic cohort includes children born between January 1, 2014, and October 31, 2022 with at least one observed encounter before age 30 days. To ensure sufficient sequence length for meaningful interpretation, we include only subjects with at least 15 diagnosis events and at most 160 events and exclude all the others. All diagnoses are encoded using ICD-10 codes. However, due to the large number of distinct ICD-10 codes, we use only the first three characters to represent each diagnosis (e.g., ICD-10 code Z00.01 is encoded as Z00) in order to group closely related codes. Our final cohort consists of 34,432 subjects, encompassing $K = 766$ event types, with an average sequence length of 75 events.

Our proposed approach aims to support clinical experts in navigating large-scale EHR datasets by identifying potential relationships between medical events. Rather than directly guiding clinical interventions, the model serves as a hypothesis-generating tool supporting hypothesis generation and a mechanistic understanding of diagnosis trajectories, aiding clinicians in pinpointing event sequences warranting further investigation through subsequent analyses or clinical trials. Additionally, our framework provides clinicians with a powerful methodology to quantitatively explore specific, clinically relevant hypotheses regarding temporal relationships between events.
\subsection{Interpret Kernel to Identifying Potential Clinical Relationships} 
To interpret the kernel function, we compute the $\int K_{i,j}(t)dt$ for each impact function forming the heatmap to show the cumulative impact from event i to event j. For this experiment we set t=1 year, and the result is shown in \cref{fig:sub3}(kernel function attached in appendix). We chose an embedding dimension of $D=3$, as experiments with higher dimensions showed inactive kernels, suggesting $D=3$ adequately captures the sparse data dynamics while maximizing interpretability.

From the heatmap, we note that input embedding dimension 2 has high impact on output embedding dimension 1; and input embedding dimension 3 has high impact on output embedding dimension 3.

To interpret these results, we first identify the specific diagnoses that load most strongly on each embedding dimension. This is done by examining the event embedding matrix: for each dimension, we rank the diagnoses according to their absolute loading values. Table \ref{table:topics} lists the selected top diagnoses for three representative topics for both the input and output embeddings.(Full top embedding loading attached in appendix) 
Having done this, we see that input embedding 2 corresponds primarily to perinatal complications; these increase subsequent likelihood of embedding dimension 1, which includes visits for otitis media and fever of unknown origin.

We then see that input embedding 3 corresponds to more serious perinatal events requiring ventilation or other critical care; which increase subsequent likelihood of embedding dimension 3, which corresponds to neurodevelopmental conditions, including speech and language concerns, and attention deficit hyperactivity disorder. This result is consistent with findings of \citet{edwards2011developmental,vargas2024effects, engelhard2023predictive}; and others, which suggest that perinatal complications confer increased likelihood of neurodevelopmental conditions.

\begin{figure}
    \centering
    \includegraphics[width=0.5\linewidth]{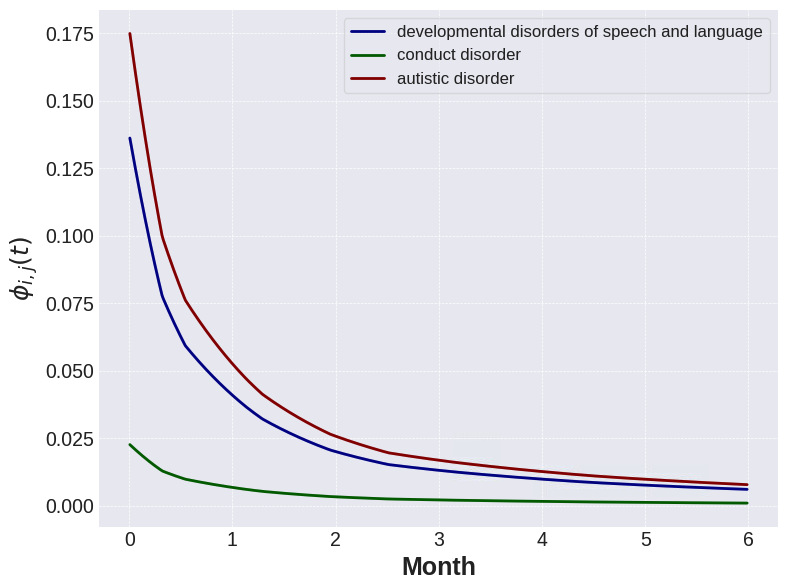}
    \caption{Recovered impact functions \(\phi_{i,j}(t)\) where $j$ is the ADHD diagnosis, $i$ is \emph{developmental disorders of speech and language} (blue line), \emph{conduct disorder} (green line), and \emph{autistic disorder} (red line), respectively}
    \label{fig:impact_functions_ADHD}
\end{figure}

\subsection{Temporal Impact Function Recovery for Actionable Insights}

Beyond uncovering high-level latent topics, the framework recovers temporal impact functions that quantify how the influence of one medical event on another evolves over time. To recover the $\phi_{i,j}$, we can simply calculate it by inserting any time interval into formula \ref{eq:method_embed}. Note that in this case study, we have two different event embeddings which correspond to the left-hand side $W$ and right-hand side $W$ in formula \ref{eq:method_embed}, respectively.  
This temporal analysis is particularly valuable when clinicians have a specific research hypothesis ---- for example, trying to understand how a prior event \(A\) influences the likelihood of a subsequent event \(B\). Our method enables recovery of the impact function \(\phi_{A,B}(t)\), which reveals how the intensity of \(A\)'s influence on \(B\) evolves over time.

For instance, prior studies and clinical observations suggest that conditions such as \emph{developmental disorders of speech and language}, \emph{conduct disorder}, and \emph{autistic disorder} may contribute to the risk and timing of an ADHD diagnosis. \citep{gu2023sex} To provide further insight to these hypotheses, the corresponding impact functions \(\phi_{i,j}(t)\) are recovered for selected event pairs in which the input event \(i\) is one of the candidate conditions and the output event \(j\) is an ADHD diagnosis. With a time window set to \(t=6\) months, the recovered impact functions provide the following insights:
Figure~\ref{fig:impact_functions_ADHD} illustrates the temporal impact functions \(\phi_{i,j}(t)\) for three potential precursors of ADHD diagnosis. Among them, \emph{autistic disorder} shows the strongest immediate influence, followed by \emph{developmental disorders of speech and language}, both decaying over time. In contrast, \emph{conduct disorder} exhibits a weaker and more prolonged effect, indicating a lower but more sustained contribution to ADHD diagnosis.  

The temporal profiles of the recovered impact functions provide several clinically actionable insights. First, they potentially help identify key windows during which specific early conditions exert the strongest influence on subsequent ADHD diagnosis, enabling clinicians to adjust their monitoring or intervention strategies in response. For example, knowing that ADHD diagnosis often follows autism diagnosis might prompt providers to consider ADHD when diagnosing autism, or alternatively to question whether changes in assessment practices may be warranted.
Second, the shape and magnitude of each function offer a quantitative basis for comparing the relative contributions of different conditions, further supporting clinical understanding. Finally, these temporal dynamics provide empirical support for clinical hypothesis in future prospective studies, offering a pathway toward validating Granger causality mechanisms suggested by the model. In the ADHD case, for example, providers might explore the impact of screening for co-occurring ADHD at the time of referral for autism evaluation, which might promote higher rates of co-incident autism and ADHD diagnoses, reducing extra visits and supporting earlier ADHD recognition. 

\section{Conclusion}
In this work, we addressed the challenge of modeling event sequences with self-reinforcing dynamics by proposing a flexible Hawkes process model that maintains interpretability. Our approach leverages a neural impact kernel in event embedding space, allowing it to capture complex event dependencies without assuming specific parametric forms, while still retaining the core interpretability of traditional Hawkes processes. By working in embedding space, our model scales to large event vocabularies and enables topic-level interpretation of event interactions.

We further introduced transformer encoder layers to contextualize event embeddings. However, experiments indicated that our flexible kernel alone sufficiently captures dynamics, often eliminating the need for additional contextualization.

Applying our method to a large-scale pediatric EHR dataset demonstrated its practical utility. In this case study, the learned event embeddings uncovered clinically meaningful diagnostic topics, while the recovered temporal impact functions \(\phi_{i,j}(t)\) provided quantitative insight into how specific early conditions—such as speech delay, conduct disorder, and autism—affect the timing and risk of future ADHD diagnosis. Thus, our model supports both exploratory analysis and validation of clinical hypotheses regarding diagnostic trajectories.

Overall, our method maintains competitive performance and interpretability, making it highly suitable for healthcare applications requiring clear insights into temporal event interactions.

\paragraph{Acknowledgment}
This work was supported the National Institute of Mental Health (K01MH127309; PI Matthew Engelhard).

\bibliography{iclr2025_conference}

\appendix
\onecolumn

\section{Training Hyperparameters}
\setlength{\abovecaptionskip}{5pt}
\setlength{\belowcaptionskip}{5pt}
\begin{table*}[htbp]
    \centering
    \caption{Hyperparameters used for the proposed method. ``\# head'' and ``\# layer'' are only applicable for contextualized embedding hyperparameters. ``MC'' stands for Monte Carlo integration, and ``NC'' stands for numerical integration.}
    \begin{tabular}{lcccccc}
        \toprule
        \textbf{Dataset} & \# head & \# layer & $D_{\text{model}}$ & Batch Size & Learning Rate & Solver \\
        \midrule
        Retweet & 1 & 1 & 3 & 64 & $1 \times 10^{-3}$ & MC \\
        Taxi & 1 & 1 & 10 & 256 & $1 \times 10^{-4}$ & MC \\
        StackOverflow & 2 & 2 & 22 & 64 & $1 \times 10^{-3}$ & MC \\
        Amazon & 2 & 2 & 16 & 64 & $1 \times 10^{-3}$ & MC \\
        MIMIC-IV & 2 & 2 & 10 & 32 & $1 \times 10^{-2}$ & NC \\
        MemeTrack & 2 & 2 & 50 & 256 & $1 \times 10^{-2}$ & NC \\
        Simulation & 1 & 1 & 3 & 128 & $1 \times 10^{-4}$ & MC \\
        \bottomrule
    \end{tabular}
    \label{tab:hyperparameters}
\end{table*}

\section{Detail description of dataset used in experiments}
\label{appendix:dataset}
\paragraph{MIMIC-IV}  The MIMIC-IV dataset contains comprehensive clinical data from patients admitted to the intensive care units (ICU) at a tertiary academic medical center in Boston. Specifically, we use procedure events, which include all medical procedures administered to patients during their ICU stay. Each procedure is time-stamped and categorized, representing different types of medical interventions. For this analysis, we included 84366 ICU stays with maximum sequence length of 240. There were $K = 159$ distinct event types \citep{johnson2023mimic}.

\paragraph{Amazon} This dataset consists of time-stamped product review events collected from Amazon users between January 2008 and October 2018. Each event includes the timestamp of the review and the category of the product being reviewed, with each category mapped to a distinct event type. In this paper, we focus on a subset of the 5,200 most active users, where each user has an average sequence length of 70, and there are $K = 16$ event types \citep{ni2019justifying}. 

\paragraph{Retweet} This dataset captures sequences of user retweets, with events classified into three types based on the size of the user’s following: ``small" (under 120 followers), ``medium" (under 1,363 followers), and ``large" (more than 1,363 followers). A subset of 5,200 active users was extracted, with an average sequence length of 70 events per user and $K = 3$ event types \citep{zhou2013learning}.

\paragraph{Taxi} This dataset logs time-stamped taxi pick-up and drop-off events across New York City’s five boroughs. Each event type is defined by the combination of the borough and whether the event is a pick-up or drop-off. This results in $K = 10$ event types. 2,000 drivers were randomly sampled with an average sequence length of 39 events \citep{whong2014foiling}.

\paragraph{StackOverflow} This dataset tracks user activities on the StackOverflow platform, specifically the awarding of badges over two years. Each event corresponds to the awarding of a badge, with $K = 22$ different badge types. For this analysis, we use a subset of 2,200 active users, each with an average sequence length of 65 \citep{jure2014snap}.

\paragraph{MemeTrack} This dataset monitors the spread of "memes" (fixed phrases) across online news articles and blogs. It records time-stamped instances of meme usage from over 1.5 million websites, where each meme defines a distinct event type. The dataset includes $K = 5000$ event types and 80000 subjects were sampled.  \citep{jure2014snap}.

Table \ref{tab:dataset-summary} summarizes the details of the datasets: 

\begin{table}[h]
\centering
\caption{Summary of datasets used in the experiments.}
\label{tab:dataset-summary}
\begin{tabular}{l r c r r r}
\toprule
 \textbf{Dataset} & \textbf{Sample Size} & \textbf{Train/Val/Test Split} & \textbf{$K$} & \textbf{Max/Avg Length} \\
\midrule
MIMIC-IV      & 84366 & 59056/12655/12655 & 159   & 240/10 \\
Amazon        & 8000  & 6454/922/1220 & 16    & 94/70 \\
Retweet       & 12000  & 9000/1500/1500 & 3     & 97/70 \\
Taxi          & 2000  & 1400/200/400 & 10    & 38/37 \\
StackOverflow & 2200  & 1400/400/400 & 22    & 101/65 \\
MemeTrack     & 80000 & 56000/12000/12000 & 5,000 & 31/3  \\
\bottomrule
\end{tabular}
\end{table}

\section{Experiment: Different model dimension vs likelihood} 

In this experiment, we evaluate the performance of models on the MIMIC-IV and Meme datasets by varying the embedding dimensions. The original event dimension of MIMIC-IV is 159, while that of Meme is significantly larger at 5,000. As expected, the reduction in log-likelihood for MIMIC-IV remains relatively small when the dimension is reduced below 10, whereas Meme experiences a more pronounced decline. This is reasonable, given that Meme’s higher original dimension suggests more potential active relationships between event types. Despite this, both datasets achieve strong log-likelihood performance in lower dimensions (10-50), indicating that many event pairs in the original impact kernel dimension are likely unrelated. This further suggests that the original impact kernel may have a sparse structure. Our method effectively recovers the impact function even in low dimensions, highlighting its ability to capture the essential relationships while reducing dimensional complexity. 

\begin{figure}[htbp]
\centering
\includegraphics[width=4.5in]{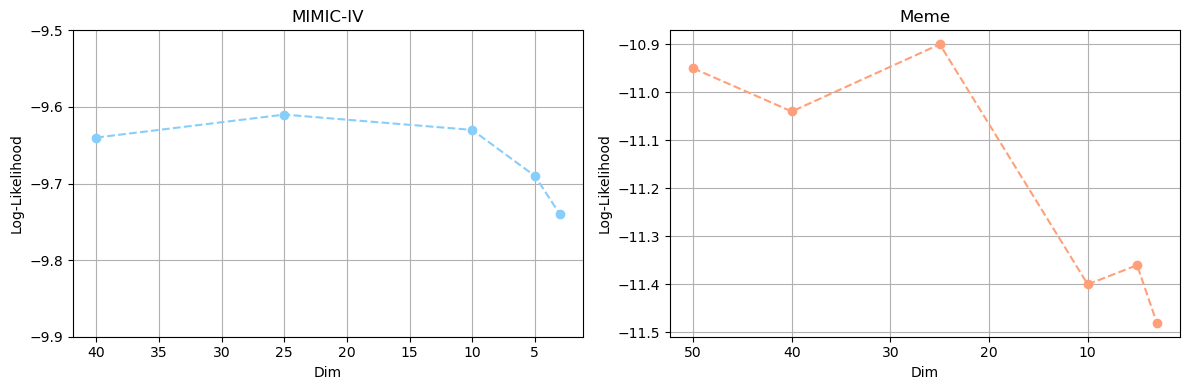}
\caption{Different dimension vs Likelihood}
\label{fig:LLvsDim}
\end{figure}

However, as shown in the figure\ref{fig:LLvsDim}, there is a noticeable drop in performance for MIMIC-IV when the dimension is set to 40. A potential reason for this decline could be the relatively long maximum sequence length in the MIMIC-IV dataset, which increases the GPU memory requirement for computation. As a result, the batch size had to be limited to 16, which might cause instability in training. Since the sequences in MIMIC-IV vary greatly in length, with some batches containing very short sequences, this variability could lead to an unstable training process, ultimately affecting the model's performance.

\newpage
\section{Kernel function of EHR case study}
\begin{figure}[h]
    \centering
    \includegraphics[width=0.9\linewidth]{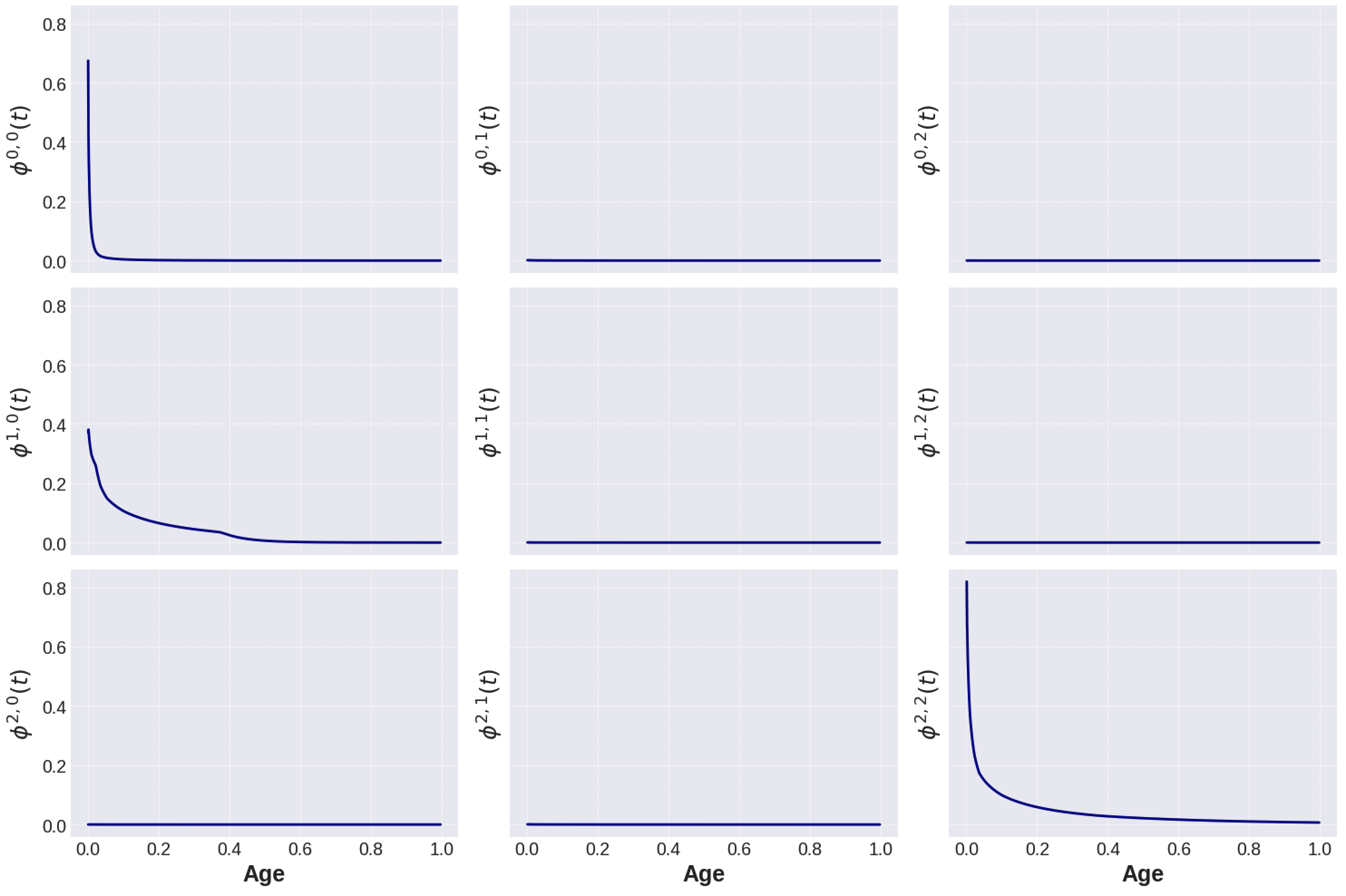}
    \caption{Kernel function of learned kernel from Duke-EHR}
    \label{fig:kernel_functions_EHR}
\end{figure}

\section{Next event prediction performance on Duke-EHR}
\begin{table*}[ht]
\centering
\caption{Next-event Prediction comparison between different models on Duke-EHR }
\resizebox{\textwidth}{!}{\begin{tabular}{l|cc|cc|cc|cc|cc}
\hline
\multirow{2}{*}{\textbf{Dataset}} 
 & \multicolumn{2}{c|}{\textbf{ENHP}} 
 & \multicolumn{2}{c|}{\textbf{AttNHP}} 
 & \multicolumn{2}{c|}{\textbf{NHP}} 
 & \multicolumn{2}{c|}{\textbf{THP}} 
 & \multicolumn{2}{c}{\textbf{SAHP}} \\
\cline{2-11}
 & \textbf{RMSE} & \textbf{error} 
 & \textbf{RMSE} & \textbf{error} 
 & \textbf{RMSE} & \textbf{error} 
 & \textbf{RMSE} & \textbf{error} 
 & \textbf{RMSE} & \textbf{error} \\
\hline
\textbf{Duke-EHR} 
 & 0.31 & 0.826
 & 0.39 & 0.820 
 & 0.39 & \textbf{0.769} 
 & 0.41 & 0.770 
 & \textbf{0.23} & 0.830 \\
\hline
\end{tabular}}
\label{tab:PredictionComp}
\end{table*}
To complement our primary analysis of interpretability, this section evaluates the predictive performance of ENHP against several state-of-the-art models on the Duke-EHR dataset. As summarized in Table \ref{tab:PredictionComp}, the results show a trade-off across metrics. Our model, ENHP, demonstrates strong performance in forecasting event timing with a highly competitive RMSE score. While its accuracy for event type prediction does not lead the benchmarks, it remains comparable to other prominent methods.

Importantly, ENHP achieves this competitive predictive performance while offering the distinct advantage of a directly interpretable kernel structure, as explored in the main case study. This balance is particularly valuable in the clinical domain, where understanding the underlying drivers and temporal relationships between medical events is often as critical as the predictive accuracy itself. Therefore, ENHP presents a compelling option for real-world EHR analysis where model transparency and hypothesis generation are paramount.
\newpage

\section{Full table of embedding from EHR}

\setlength{\abovecaptionskip}{-12pt}
\setlength{\belowcaptionskip}{-12pt}
\begin{table}
\floatconts
  {tab:topics} 
  {\caption{Event Embedding for Topic Discovery on EHR, rank by embedding loading value(in the bracket)}} 
  {\footnotesize 
  \resizebox{\textwidth}{!}{
    \begin{tabular}{cp{4.8cm}p{4.8cm}p{4.8cm}} 
      \toprule
      \multicolumn{4}{c}{\textbf{Event Embedding for Input Events}} \\ 
      \midrule
      \textbf{Rank} & \textbf{Topic 1} & \textbf{Topic 2} & \textbf{Topic 3} \\ 
      \midrule
      0 & Neonatal jaundice from other and unspecified causes (2.5991) & Acne (1.2489) & Dependence on enabling machines and devices, NEC (1.8859) \\
      1 & Feeding problems of newborn (1.8985) & Newborn affected by maternal complications of pregnancy (1.0613) & Respiratory distress of newborn (1.4855) \\
      2 & Unspecified jaundice (1.8888) & NB aff by matern cond that may be unrelated to present preg (1.0554) & Encounter for attention to artificial openings (0.9735) \\
      3 & Other conditions of integument specific to newborn (1.2797) & Other congenital infectious and parasitic diseases (0.8658) & Cardiovascular disorders originating in the perinatal period (0.9559) \\
      4 & Umbilical hemorrhage of newborn (1.2588) & Newborn affected by other comp of labor and delivery (0.8150) & Transitory disord of carbohydrate metab specific to newborn (0.9012) \\
      5 & Other conditions originating in the perinatal period (1.2495) & Seborrheic dermatitis (0.7765) & Neonatal aspiration (0.8783) \\
      6 & Disorders of porphyrin and bilirubin metabolism (1.2250) & Maternal care for malpresentation of fetus (0.6787) & Other venous embolism and thrombosis (0.7190) \\
      7 & Other infections specific to the perinatal period (0.9996) & Hemangioma and lymphangioma, any site (0.6465) & Shock, not elsewhere classified (0.6834) \\
      8 & Newborn affected by comp of placenta, cord and membranes (0.9810) & Encounter for administrative examination (0.5881) & Nontraumatic intracerebral hemorrhage (0.6803) \\
      9 & Oth respiratory conditions origin in the perinatal period (0.9652) & Umbilical hernia (0.5650) & Other problems with newborn (0.6537) \\
      \midrule
      \multicolumn{4}{c}{\textbf{Event Embedding for Output Events}} \\ 
      \midrule
      \textbf{Rank} & \textbf{Topic 1} & \textbf{Topic 2} & \textbf{Topic 3} \\ 
      \midrule
      0 & Encntr for general exam w/o complaint, susp or reprtd dx (5.1050) & Encounter for immunization (0.9134) & Specific developmental disorders of speech and language (2.0099) \\
      1 & Encounter for immunization (2.4261) & Suppurative and unspecified otitis media (0.1199) & Disord of NB related to short gest and low birth weight, NEC (1.2127) \\
      2 & Suppurative and unspecified otitis media (1.1593) & Persons encntr hlth serv for spec proc \& trtmt, not crd out (0.0949) & Personal risk factors, not elsewhere classified (1.0187) \\
      3 & Acute upper resp infections of multiple and unsp sites (1.0799) & Fever of other and unknown origin (0.0890) & Congenital malformations of cardiac septa (1.0022) \\
      4 & Fever of other and unknown origin (0.7903) & Contact w and (suspected) exposure to communicable diseases (0.0872) & Lack of expected normal physiol dev in childhood and adults (0.8759) \\
      5 & Feeding problems of newborn (0.6888) & Cough (0.0730) & Symptoms and signs concerning food and fluid intake (0.6744) \\
      6 & Persons encntr hlth serv for spec proc \& trtmt, not crd out (0.6069) & Feeding problems of newborn (0.0616) & Other disorders of muscle (0.6726) \\
      7 & Cough (0.5554) & Oth symptoms and signs involving the circ and resp sys (0.0578) & Conductive and sensorineural hearing loss (0.6637) \\
      8 & Contact w and (suspected) exposure to communicable diseases (0.4490) & Abnormalities of breathing (0.0560) & Oth symptoms and signs involving the nervous and ms systems (0.5820) \\
      9 & Atopic dermatitis (0.4074) & Atopic dermatitis (0.0548) & Attention-deficit hyperactivity disorders (0.5224) \\
      \bottomrule
    \end{tabular}}
  }
  \label{table:topics_full}
\end{table}

\newpage

\section{Interpretation of Kernel from MIMIC}
\begin{table*}[h]
\floatconts
  {tab:event-embedding-mimic} %
  {\caption{Event Embedding for Topic Discovery on MIMIC-IV}} %
  {\footnotesize %
    \begin{tabular}{c|p{4cm}|p{4cm}|p{4cm}} %
      \toprule
      \multicolumn{4}{c}{\textbf{Event Embedding for Input Events}} \\ 
      \midrule
      \textbf{Rank} & \textbf{Input 1} & \textbf{Input 2} & \textbf{Input 3} \\ 
      \midrule
      0 & Extubation & Nasal Swab & 14 Gauge \\ 
      1 & Presep Catheter & Indwelling Port & Indwelling Port \\ 
      2 & TLS Clearance & Indwelling Port (PortaCath) & 16 Gauge \\ 
      3 & Esophogeal Balloon & Foley Catheter & Intraosseous Device \\ 
      4 & Chest Opened & ERCP (Travel to) & Presep Catheter \\ 
      5 & Rectal Swab & Presep Catheter & MAC \\ 
      6 & Venogram & Transfer Intercampus by Ambulance & IABP line \\ 
      7 & Peritoneal Dialysis & Pheresis Catheter & 18 Gauge \\ 
      8 & Dialysis - CRRT & TLS Clearance & Transfer Intercampus by Ambulance \\ 
      9 & EEG & Endoscopy & RIC \\ 
      \midrule
      \multicolumn{4}{c}{\textbf{Event Embedding for Output Events}} \\ 
      \midrule
      \textbf{Rank} & \textbf{Output 1} & \textbf{Output 2} & \textbf{Output 3} \\ 
      \midrule
      0 & Chest X-Ray & Invasive Ventilation & 20 Gauge \\ 
      1 & EKG & OR Received & 18 Gauge \\ 
      2 & Blood Cultured & Intubation & Arterial Line \\ 
      3 & Family updated by RN & Foley Catheter & 16 Gauge \\ 
      4 & Portable Chest X-Ray & OR Sent & Multi Lumen \\ 
      5 & Urine Culture & Multi Lumen & 22 Gauge \\ 
      6 & Family updated by MD & Arterial Line & Foley Catheter \\ 
      7 & CT scan & Nasal Swab & PA Catheter \\ 
      8 & Transthoracic Echo & Chest X-Ray & Cordis/Introducer \\ 
      9 & Ultrasound & Chest Tube Removed & PICC Line \\ 
      \bottomrule
    \end{tabular}
  }
  \label{table:interpretation}
\end{table*}

\begin{figure}[htbp]
\floatconts
  {fig:mimic-kernel} %
  {\caption{Heatmap of different time of integration with regard to kernel function on MIMIC-IV}} %
  {%
    \subfigure[Integrate from 0 to 2 hour]{\label{fig:mimicsub1}%
      \includegraphics[width=0.4\linewidth]{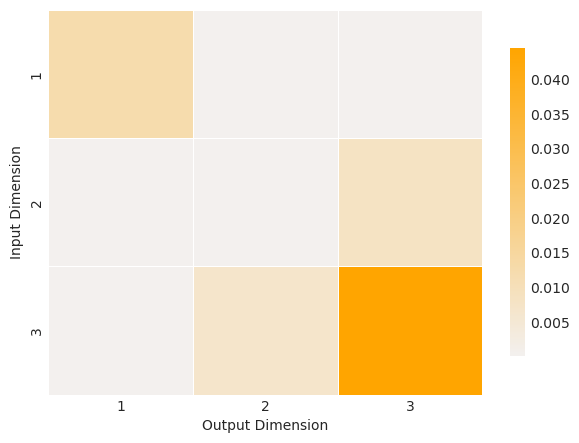}}%
    \hfill
    \subfigure[Integrate from 0 to 100 hour]{\label{fig:mimicsub2}%
      \includegraphics[width=0.4\linewidth]{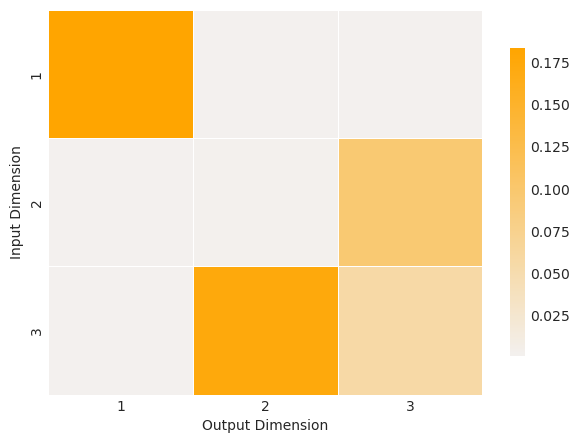}}%
  }
{\label{fig:mimic-kernel}}
\end{figure}

Results on MIMIC-IV allow us to illustrate how our method facilitates interpretability. We begin by identifying specific procedures in the dataset that load most strongly on each embedding dimension (see Table \ref{table:interpretation}). The input embeddings reveal distinct clinical workflow patterns through their dominant procedural clusters. Input 1 captures post-interventional transitions, exemplified by procedures like Extubation and Peritoneal Dialysis, which collectively reflect patient stabilization phases after critical interventions. Input 2 emphasizes diagnostic and device maintenance protocols, including Nasal Swab and Indwelling Port management, aligning with infection surveillance or catheter-associated complication prevention. Input 3 prioritizes technical specifications, such as 14/16/18 Gauge device selection and Intraosseous Device use, suggesting scenario-driven decisions balancing urgency and precision. Correspondingly, output embeddings map to downstream clinical actions: Output 1 integrates imaging (Chest X-Ray) and cardiac monitoring (EKG) for post-procedural verification, Output 2 links procedural escalation (Invasive Ventilation, OR Received) with line placement (Arterial Line), and Output 3 codifies technical specificity in interventional radiology (20/18 Gauge, Cordis/Introducer), reflecting hierarchical decision-making from coarse to fine instrumentation.  

The cross-dimensional relationships (Figure \ref{fig:mimic-kernel}) expose clinically meaningful workflows. Here, when computing $\int K_{i,j}(t)dt$ for each impact function, we consider two different $t$. Figure \ref{fig:mimicsub1} shows short-term impact, where $t=2$hours and Figure \ref{fig:mimicsub2} shows long-term impact, where $t=100$hours. For short-term impact, the major impact is from input dimension 3 to out dimension 3.  Input 3’s focus on metric-driven instrumentation (14/16 Gauge, Intraosseous Device) directly maps to Output dim3’s prioritization of caliber-defined tools (20/18 Gauge, Cordis/Introducer). This suggests: A hierarchical decision logic: Coarse-gauge devices in Input dim3 (e.g., 14G for rapid fluid resuscitation) may precede finer-gauge outputs (e.g., 20G for targeted drug delivery), reflecting escalation from urgent stabilization to precision therapy. Also : Intraosseous access (Input dim3) often necessitates subsequent central line placement (Output dim3’s Cordis/Introducer), adhering to trauma resuscitation protocols where temporary access transitions to definitive vascular support. On the other hand, the long-term impact, the major impact is from input dimension 1 to output dimension 1 and input dimension 3 to output dimension 2. The first one reflect Post-procedural interventions (e.g., TLS Clearance, Extubation) correlate strongly with imaging workflows (Chest X-Ray, Portable X-Ray), likely reflecting post-intervention verification (e.g., confirming endotracheal tube placement or lung re-expansion). The second impact(dim 3 to 2) shows standardized device inputs (14 Gauge) associate with line placement (Arterial Line, Multi Lumen), reflecting resource optimization in critical care (e.g., using larger-bore devices for hemodynamic monitoring). 

This shows that our method could capture the impact machinisim in different time window which provide enough flexibility in real-world application. On the other hand, while the low-dimensional embedding (dim=3) preserves interpretability, it conflates mechanistically distinct sub-processes. For example, Input 1 merges respiratory weaning (Extubation) with renal support (Peritoneal Dialysis), obscuring specialty-specific pathways, while Output 2 combines surgical preparation (OR Received) with ventilation management, masking phase-specific logic. This simplification may artificially inflate associations between unrelated workflows (e.g., linking intraosseous access to PICC line placement via gauge metrics). Future work should explore higher-dimensional embeddings to isolate granular topics, such as differentiating emergency vascular access from elective line selection or disentangling adult vs. pediatric device specifications. Such refinements could enhance clinical utility by aligning computational topics with domain-specific decision hierarchies.

\end{document}